\documentclass[letterpaper,times]{IONconf}
\usepackage{url}
\usepackage{natbib}
\usepackage{graphicx}  % figures alignment
\usepackage{booktabs}  % lines for tables
\usepackage{amsmath}  % for specific commands in math mode
\usepackage{amssymb}  % for specific symbols in math mode
\usepackage{algorithm}  % for pseudo-code
\usepackage{algpseudocode}  % for pseudo-code
\usepackage{subfig}     % for side-by-side figures
\usepackage[title]{appendix}    % To have appendix titles
\usepackage{float}  % forcing figure placement

\usepackage{xcolor}         % colors

\DeclareMathOperator*{\Cnt}{\mathbf{Count}}

\usepackage{hyperref}  % Davide replaced with this to see links
\title{GNSS Positioning using Cost Function Regulated Multilateration and Graph Neural Networks}

\author{
    Amir~Jalalirad, Davide~Belli, Bence~Major, \textit{Qualcomm~AI~Research}% <- this '%' removes a trailing whitespace
    \footnotemark[1]
    \vspace{1mm} \\%
    Songwon~Jee, Himanshu~Shah, Will~Morrison, \textit{Qualcomm~Technologies,~Inc.}% <- this '%' removes a trailing whitespace
    }

\begin{document}

\maketitle
\footnotetext[1]{Qualcomm AI Research is an initiative of Qualcomm Technologies, Inc.}

% ######################################################################################################
% ######################################## BIBLIOGRAPHY ################################################
% ######################################################################################################
% biography section. The * indicates a section excluded from numbering.
\section*{biography}

% Biographies are defined as follows:
% \biography{Author name}{author biography text}

\biography{Amir Jalalirad}{is a Staff Machine Learning Researcher at Qualcomm AI Research. He obtained his Ph.D. in Electrical Engineering from Eindhoven University of Technology in 2016. His research interests include applications of deep learning in positioning, navigation and RF signal processing systems.}

\biography{Davide Belli}{received his M.S. degree in Artificial Intelligence from the University of Amsterdam in 2019. He is currently a Senior Machine Learning Researcher at Qualcomm AI Research. His research interests include deep learning for the visual and RF domain, model personalization, and graph representation learning.}

\biography{Bence Major}{is a Staff Engineer at Qualcomm AI Research, leading a research team in the use of artificial intelligence for RF sensing and positioning. His research work focuses on non-visual sensory data, such as radar, ultrasound, and wireless signals. He received his M.S. degree in Computer Science from the Budapest University of Technology and Economics.}

\biography{Songwon Jee}{received his M.S. degree in Electrical Engineering from Stanford University in 2016. He is currently a Senior Staff Engineer in Location Technology Team at Qualcomm Technology Inc. His research interests include the application of deep learning for location technology involving GNSS, sensors, and wireless technologies.}

\biography{Himanshu Shah}{received his M.S. and Ph.D. degrees in Electrical Engineering from Arizona State University in 2004 and 2009 respectively. He has been involved in the research, design and development of hybrid location technologies for wireless devices since 2010. He is currently a Principal Engineer at Qualcomm Technologies Inc.}

\biography{Will Morrison}{is a Director of Engineering on the Location Team at Qualcomm. He has 25 years of experience in research and development of hybrid positioning technology for mass deployment in mobile, automotive and IOT devices. During his tenure at Qualcomm, Will has held central roles in the development of fused multi-constellation GNSS, MEMS and wireless technology navigation solutions. He holds B.S. and M.S. degrees in Mechanical Engineering (Control Systems) from Columbia University and University of California at Berkeley.}

% ######################################################################################################
% ######################################## ABSTRACT ####################################################
% ######################################################################################################
% The Abstract. The * indicates a section excluded from numbering.
\section*{Abstract}

In urban environments, where line-of-sight signals from GNSS satellites are frequently blocked by high-rise objects, GNSS receivers are subject to large errors in measuring satellite ranges. Heuristic methods are commonly used to estimate these errors and reduce the impact of noisy measurements on localization accuracy. In our work, we replace these error estimation heuristics with a deep learning model based on Graph Neural Networks. Additionally, by analyzing the cost function of the multilateration process, we derive an optimal method to utilize the estimated errors. Our approach guarantees that the multilateration converges to the receiver's location as the error estimation accuracy increases. We evaluate our solution on a real-world dataset containing more than 100k GNSS epochs, collected from multiple cities with diverse characteristics. The empirical results show improvements from 40\% to 80\% in the horizontal localization error against recent deep learning baselines as well as classical localization approaches.

% ######################################################################################################
% ######################################## INTRODUCTION ################################################
% ######################################################################################################
% The introduction. Section numbering starts here.
\section{Introduction}

In Global Navigation Satellite System (GNSS) positioning, the receiver is localized with respect to several satellites in the sky. Each satellite sends out a unique signal, and the receiver estimates its distance to each satellite, known as a pseudo-range, by measuring the signal propagation time. The set of distances from the visible satellites are used in multilateration to find a location on Earth. This process becomes challenging in dense urban environments where tall buildings block the direct path from satellites to the receiver, causing non-line-of-sight (NLOS) and multi-path conditions, and leading to increased errors in distance estimation.

In this work, we focus on the “first fix” scenario, in which an initial positioning estimate is requested for the first time without having access to any previous measurements or location. Regularly, methods addressing the first fix problem find the location at which the computed distances to the satellites best match the measured pseudo-ranges. This is done by first defining an objective function that associates a cost to each possible location (and receiver clock bias) and then using an optimization method to find a point where the cost is minimal. The most prevalent objective function is the weighted sum of squared residuals, where residual is defined as the difference between the computed distance to a satellite at a given location and the measured pseudo-range \citep{noureldin2012fundamentals, betz2021fundamentals}. Commonly, a variant of the Newton method often referred to as Weighted Least Squares (WLS) is used iteratively to optimize the cost.

Many of such solutions estimate pseudo-range measurement errors based on some satellite or signal property like satellite elevation or carrier power to noise density ratio (C/N\textsubscript{0}) \citep{groves2013height, won2012weighted}. Then, these estimates are heuristically mapped to weights for WLS, without guarantees that perfect estimates yield the true position. This introduces two sources of inaccuracy: the first being inaccurate estimation of errors, and the second being the suboptimal way of using estimated errors in the optimization process.

We aim to reduce the localization inaccuracy introduced by the first source by designing an expressive graph neural network to estimate measurement errors. In addition, we introduce a measurement selection unit that reduces the set of used measurements to a more robust subset.

We then remove the second source of inaccuracy by introducing a cost function regulation process. This process guarantees that perfect measurement error estimates force the truth location to have minimal cost. Thus, the following optimization process is guaranteed to yield the truth location. 

The main contributions of our research are summarized as follows:
\begin{itemize}
    \item We analyze the optimization process involved in multilateration-based localization. We show how the cost function for this optimization can be adjusted for optimal localization using measurement errors.
    \item We propose using a graph neural network model to jointly process information across GNSS measurements in an epoch and produce reliable error estimates per measurement.
    \item We introduce an adaptive GNSS measurement selection algorithm that further improves receiver positioning in case of suboptimal error estimation.
\end{itemize}

% ######################################################################################################
% ######################################## RELATED WORK ################################################
% ######################################################################################################
\section{Related Work}

Popular localization solutions adjust the optimization objective function so that the least accurate pseudo-range measurements are generally weighted less compared to the more accurate ones. Extended research has been carried out to determine a weighting scheme that improves the error in locations predicted by WLS. For instance, signals that refract or reflect off surfaces on their paths are usually relatively weak and contribute to large errors in pseudo-range measurements. \citet{hartinger1999variances} and \citet{groves2013height} use the C/N\textsubscript{0} of the received signal as an indicator of measurement errors and design weighting schemes based on it. As the signal from a low-elevation satellite is more likely to be disturbed by high-rise objects on the earth’s surface, researchers found satellite elevation to be another relevant factor for designing weighting schemes. \citet{won2012weighted} fit a first-order exponential function of the satellite elevation to the training data to model measurement weights. \citet{tabatabaei2017reliable} propose instead a fuzzy system to weight measurements according to a set of predefined rules based on satellite elevation angle and Dilution of Precision. \citet{akram2018gnss} use well-established mathematical functions of the residuals to iteratively re-weight measurements in WLS. In our work, we prove that the set of optimal weights for the WLS algorithm is not unique and the weights that result in an accurate location prediction are tractable using pseudo-range measurement errors.

In the last decade, neural networks have been used extensively to model complex relations and patterns in GNSS data and improve different steps of GNSS localization. In \citep{suzuki2021nlos}, the authors use a Multi-Layer Perceptron (MLP) to detect LOS/NLOS using the receiver’s auto-correlation signal. In other work \citep{kanhere2022improving, siemuri2021improving}, neural networks have been trained to directly produce location corrections with respect to an input anchor point. While some researchers address the first fix problem and avoid using time correlations in the input data \citep{caparra2021machine}, others have used neural networks in combination with Kalman filters to accurately locate \citep{mohanty2022learning} and track the receiver \citep{guo2021novel, han2021precise, gupta2022designing}. Finally, neural networks have also been used to estimate pseudo-range measurement errors. In \citep{zhang2021prediction}, the authors represent the measurements in an epoch as a sequence of feature vectors and use a Long Short-Term Memory (LSTM) \citep{hochreiter1997long} unit to process them and predict visibility and estimate pseudo-range errors. While the sequential representation allows handling the varying number of measurements across epochs, it also requires the definition of an ordering over the sequence elements. Since there is no obvious way to define an ordering over measurements, the authors randomly permute the measurement order when training the neural network. This is a suboptimal choice since the model must learn at the same time to be order-invariant and to predict the correct labels \citep{wu2020comprehensive, bronstein2021geometric}. As it has been extensively studied in recent deep learning literature, increased model performance may be reached by designing neural network architectures that are invariant or equivariant to symmetries in the data \citep{you2018graphrnn, zaheer2017deep}. 

We propose to represent the measurements in an epoch as a complete graph, by defining the set of nodes as the measurements and the edges as the angular proximity between each pair of satellites in the observable hemisphere. Since this representation is order-invariant, we remove the need for an arbitrary ordering among measurements, and we can use the family of deep learning models called Graph Neural Networks (GNNs) \citep{scarselli2008graph, kipf2016semi, xu2018powerful, zhou2020graph} to propagate information across the epoch’s measurement. To the best of our knowledge, the only existing work adopting GNNs for GNSS positioning \citep{mohanty2022learning} defines the graph’s structure by connecting measurements with the same constellations and frequencies. In this work, we instead propose to define the graph connectivity based on the location of the satellites in the visible hemisphere. By representing the geometrical proximity among satellites, the GNN can learn to detect and correct inconsistencies in the measured pseudo-ranges by comparing multiple observations from the same sky region.

\begin{figure}[htb]
\centering
\includegraphics[width=\textwidth]{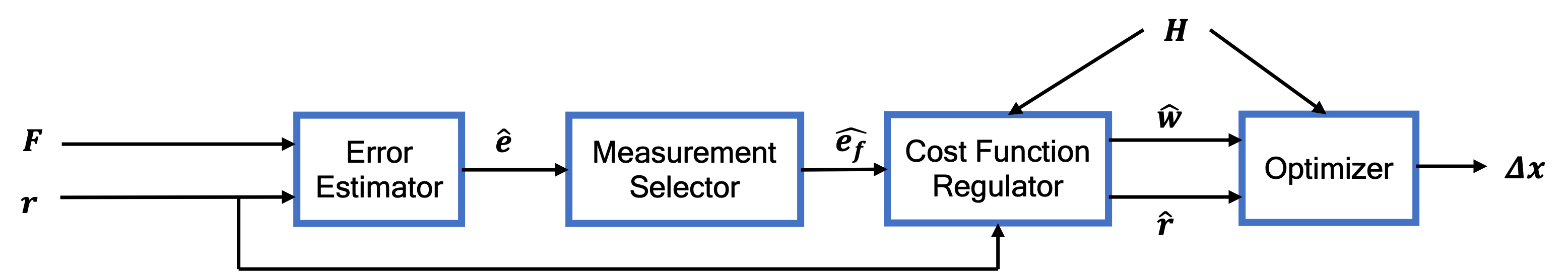}
\caption{Pipeline of the proposed solution. $\mathbf{F}$: input features, $\mathbf{r}$: residuals at a guess location, $\mathbf{\hat{e}}$: error estimates, $\mathbf{\hat{e}_f}$: selected error estimates, $\mathbf{H}$: geometry matrix, $\mathbf{\hat{w}}$: adjusted weights, $\mathbf{\hat{r}}$: adjusted residuals, $\mathbf{\Delta x}$: predicted displacement from guess location.}
\label{fig:pipeline}
\end{figure}

% ######################################################################################################
% ######################################## METHOD ######################################################
% ######################################################################################################
\section{Method}

Our proposed method for GNSS localization consists of four main components. The first component is an error estimation model which processes a set of features extracted from the GNSS signals and the satellite data and predicts a set of pseudo-range measurement errors. The second is a measurement selection component, where pseudo-range measurements are filtered based on the distribution of the predicted errors. The third component is the Cost Function Regulator block. In this component, the pseudo-range measurements and their weights are adjusted so that the truth location has minimal cost in the optimization. Finally, we have an optimization block in which WLS uses the adjusted measurements and weights to output the location prediction. To motivate the design of our Error Estimator and Measurement Selector components, we first need to establish the purpose and structure of the Cost Function Regulator.

\subsection{Cost Function Regulator}

We consider the weighted sum of squared (pre-fix) residuals\footnotemark[2] as the cost function of the optimization process for localization. At training time, when we have access to the truth location, we discover and enforce the constraints that pseudo-range measurements and their weights need to satisfy, for the cost function to have a minimum at the truth location. At inference time, the same constraints on pseudo-range measurements and their weights are enforced in the Cost Function Regulator block.

\footnotetext[2] {Pre-fix residuals, sometimes called innovations, are the residuals computed before updating the localization system's state with new measurements. In this work, we simply refer to this concept by the term "residual".}

Let us begin by defining localization as an optimization problem. At each epoch, the receiver at position $\left(x_{t}, y_{t}, z_{t}\right)$ with receiver clock bias towards satellite's clock $\delta t_{t}$ (in meters), measures $n$ pseudo-ranges. After correcting for estimated satellite clock offsets and ionospheric and tropospheric delays, the measured pseudo-ranges can be written as
\begin{equation} 
\label{eq:meas}
m^{\left(i\right)} = \sqrt{\left(x^{\left(i\right)}_s-x_t\right)^2 + \left(y^{\left(i\right)}_s-y_t\right)^2 + \left(z^{\left(i\right)}_s-z_t\right)^2} + \delta t_t + e^{\left(i\right)} ,
\end{equation}
where the i\textsuperscript{th} satellite position is $\left(x^{\left(i\right)}_s, y^{\left(i\right)}_s, z^{\left(i\right)}_s\right)$ and $e^{\left(i\right)}$ represents the measurement errors remaining after all corrections. The goal is to find the set of unknowns, or simply the ``location'', $\mathbf{x}=\left(x, y, z, \delta t\right)$, for which the computed pseudo-ranges
\begin{equation} 
\label{eq:pseudorange}
\rho^{\left(i\right)}_{\mathbf{x}} = \sqrt{\left(x^{\left(i\right)}_s-x\right)^2 + \left(y^{\left(i\right)}_s-y\right)^2 + \left(z^{\left(i\right)}_s-z\right)^2} + \delta t ,
\end{equation}
best match the measured pseudo-ranges $m^{\left(i\right)}$. To that end, a cost function is defined to assign a cost to each location. As mentioned earlier, one of the most prevalent cost functions for this purpose is the weighted sum of squared residuals:
\begin{equation} 
\label{eq:wssr}
S_{\mathbf{x}} = \sum_{i=1}^n w^{\left(i\right)} \left(\rho^{\left(i\right)}_{\mathbf{x}}-m^{\left(i\right)}\right)^2 = \sum_{i=1}^n w^{\left(i\right)} {\left(r^{\left(i\right)}_{\mathbf{x}}\right)}^2,
\end{equation}
where residual $r^{\left(i\right)}_{\mathbf{x}}$ is the difference between the measured and computed pseudo-range at location $\mathbf{x}$. The weights $w^{\left(i\right)}$ are used to adjust the importance of one measurement against the others. It is a common practice to linearize Eq.~\ref{eq:pseudorange} around a guess location $\mathbf{x}_0$ and convert the cost function in Eq.~\ref{eq:wssr} to a system of linear equations. Then, an optimization process, such as WLS, is used to solve the linear system of equations and find the location corresponding to a local minimum. Residuals are computed at the solution location and a new system of equations is defined. This process is repeated iteratively until convergence \citep{betz2021fundamentals}. The solution to this system of equations can be far from the receiver location depending on the measurement noise. In our proposed approach, we ensure that the ground truth location indeed solves this system of equations.

We start from the fact that at the local minima, the partial derivatives of the cost function are zero. Taking the partial derivatives of Eq.~\ref{eq:wssr} with respect to $\mathbf{x}$ and setting to zero, we have
\begin{equation} 
\label{eq:partial}
\frac{\partial S_{\mathbf{x}}}{\partial \mathbf{x}} = \sum_{i=1}^n 2 w^{\left(i\right)} \frac{\partial \rho^{\left(i\right)}_{\mathbf{x}}}{\partial \mathbf{x}}\left(\rho^{\left(i\right)}_{\mathbf{x}}-m^{\left(i\right)}\right)=0 .
\end{equation}
Any local minimum of the cost function will satisfy this equation. As we want the ground truth location to be a minimum cost point, the vector of residuals at the truth location must satisfy this equation (though this condition is not sufficient, we empirically found that our method is not affected by other local minima). Using $\mathbf{x}=\mathbf{x_t}$ in Eq.~\ref{eq:pseudorange} and replacing Eq.~\ref{eq:meas} and Eq.~\ref{eq:pseudorange} in Eq.~\ref{eq:partial}, we have
\begin{equation} 
\label{eq:partialatgt}
\frac{\partial S_{\mathbf{x}}}{\partial \mathbf{x}} \Big|_{\mathbf{x}=\mathbf{x_t}}= \sum_{i=1}^n 2 w^{\left(i\right)} \frac{\partial \rho^{\left(i\right)}_{\mathbf{x}}}{\partial \mathbf{x}}\Big|_{\mathbf{x}=\mathbf{x_t}} \left(e^{\left(i\right)}\right)=0 .
\end{equation}
Note that the cost function $S_{\mathbf{x}}$ has $n$ scalar terms and $\mathbf{x}$ is a column vector of 4 components. So we have 4 first-order derivatives per each term in the cost function. Knowing that the elements of the Jacobian $\nabla_\mathbf{x} \mathbf{\rho}$ correspond to elements of the geometry matrix ($\frac{\partial \rho^{\left(i\right)}_{\mathbf{x}} } {\partial x_j} = h_{i,j}$), we can rearrange Eq.~\ref{eq:partialatgt} to the matrix form
\begin{equation} 
\label{eq:hwe}
\mathbf{H^TWe}=\mathbf{0}.
\end{equation}
where $\mathbf{W}$ is the diagonal matrix of weights $w^{\left(i\right)}$, $\mathbf{H^T}$ is the transpose of geometry matrix, and $\mathbf{e}$ represents the column vector of measurement errors $e^{\left(i\right)}$ in the epoch. 

Since the geometry matrix is dictated by the position of the satellites, we can regulate the cost function by adjusting the two remaining variables, measurement errors $\mathbf{e}$ and weights $\mathbf{W}$, so that Eq.~\ref{eq:hwe} holds.

In the first variant of our solution, we regard Eq.~\ref{eq:hwe} as the matrix form of an underdetermined system of equations with weights as unknowns. By rearranging the terms in each of the 4 equations in Eq.~\ref{eq:hwe},
\begin{equation} 
\label{eq:hew}
\mathbf{H_e^T}\mathbf{w}=\mathbf{0}.
\end{equation}
where $\mathbf{H_e^T}$ is the column-wise multiplication of $\mathbf{H^T}$ by $\mathbf{e}$ and $\mathbf{w}$ represents column vector of weights. Using any point in the null space of $\mathbf{H_e^T}$ as the set of weights for WLS, results in the ground truth user location being a local minimum of the cost function. Since a point (set of weights) in this null space is defined by $n-4$ free parameters and 4 constrained parameters, the set of solutions to Eq.~\ref{eq:hew} construct an ($n-4$)-dimensional hyper-plane in the $n$-dimensional space, proving that the optimal set of weights for WLS is not unique. Furthermore, this shows that any estimation method that assigns weights to a GNSS measurement in isolation from the rest of the measurements in the epoch is prone to failure. In our solution, we compute the optimal weights by computing the null space of the $\mathbf{H_e^T}$ matrix. Then, we pick an arbitrary point in the null space and use it as the weights in the cost function. We name this variant of our solution ``Weight Regulation''.

In the second variant of our solution, we make Eq.~\ref{eq:hwe} hold by ensuring that the residuals at the ground truth location are zero. We achieve this by adding a correction to each pseudo-range measurement and therefore to each residual, so that
\begin{equation} 
\label{eq:hwec}
\mathbf{H^TW(e+c)}=\mathbf{0}.
\end{equation}
The equation above holds if
\begin{equation} 
\label{eq:reskernel}
\mathbf{c}=\mathbf{-e} + \mathbf{v}; \:\:\: \mathbf{v} \in \mathbf{Ker\{H^TW\}},
\end{equation}
where $\mathbf{Ker\{L\}}$ refers to the kernel space of the linear map $\mathbf{L}$. Since any vector $\mathbf{c}$ from Eq.~\ref{eq:reskernel} leads to minimum cost at the ground truth location, the set of used weights is an arbitrary choice, and we use an identity matrix. Also, as the vector of zeros is a point in the kernel space of $\mathbf{H^TW}$, it suffices to have the error vector $\mathbf{e}$ for finding an optimal solution. We name this variant of our solution ``Measurement Regulation''.

Both variants of our solution guarantee that the user localization error will be zero given the vector of measurement errors. However, since these errors are only available at training time, we need a model to estimate them at the time of inference.

\begin{figure}[htb]
\centering
\includegraphics[width=\textwidth]{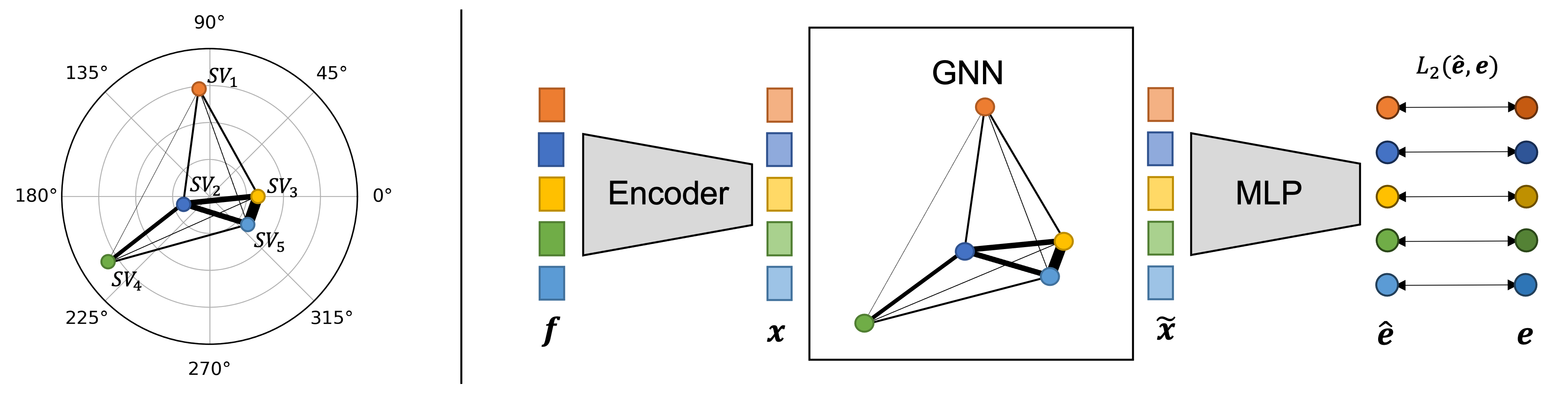}
\caption{On the left, we show with an example how we define a graph over 5 measurements in an epoch. The thickness of the edges represents the value of the angular proximity between each pair of satellites. On the right, we draw a simplified diagram for the Error Estimator neural network. First, measurement features $\mathbf{f}$ are independently encoded into latent vectors $\mathbf{x}$. Then, the Graph Neural Network propagates information across all measurements in the epoch, and an MLP outputs the estimated residuals at the ground truth location $\mathbf{\hat{e}}$. The $L_2$ loss is used for the supervised training given the ground truth residual labels $\mathbf{e}$.}
\label{fig:nn}
\end{figure}

\subsection{Estimation of Measurement Error}

We propose using a neural network to estimate the GNSS measurement errors which are later used by the Cost Function Regulator block. Since we are considering the first fix scenario, the inputs to the neural network will be extracted from a single GNSS epoch. Details on the selection and processing of the neural network inputs are included in Sec.~\ref{subsec:exp_dataset}.

We represent each epoch as a graph $G = \{\mathbf{X}, \mathbf{A}\}$, with node features $\mathbf{X} = \{ \mathbf{x}^{\left(i\right)} \in \mathbb{R}^{D}; \: 1 \leq i \leq n \}$ and adjacency matrix $\mathbf{A} = \{ a_{ij} \in [0, 1]; \: 1 \leq i, j \leq n \}$ where $n$ is the number of measurements in the epoch and $D$ is the size of the feature set for each measurement. The edges, which describe the similarity between pairs of nodes, have the assigned value of angular proximity between the two satellites. We choose the proximity between satellites because we want the neural network to be able to compare multiple observations from the same sky region, which can be useful to detect and correct inconsistencies in the measured pseudo-ranges. In Fig.~\ref{fig:nn} (left) we show a sample graph for an epoch with measurements from 5 satellites. The thickness of the edges in the graph represents the value of the angular proximity between each pair of satellites, with satellites close to each other having higher edge values, and farther satellites having lower edge values.

We employ a Graph Neural Network to process the measurements and propagate information over the graph. We choose GNNs because they have the benefit of being invariant to the order of input measurements and can process a varying number of measurements in each epoch. We implement this as a 2-layer GraphSAGE \citep{hamilton2017inductive} neural network with weighted mean as the aggregation function. As we discuss in Sec.~\ref{subsec:results}, we find this architecture to be optimal with respect to other types of GNNs. 

The overall architecture for Error Estimator neural network is shown in Fig.~\ref{fig:nn} (right). An MLP encoder $\phi_e$ is used to independently project the input features $\mathbf{f}^{\left(i\right)}$ for each of the $n$ measurements as $\mathbf{x}^{\left(i\right)} = \phi_e(\mathbf{f}^{\left(i\right)})$. The information in the latent features $\mathbf{x}^{\left(i\right)}$ is then propagated through the graph neural network $\textrm{GNN}_\theta$ according to the graph structure defined by the adjacency matrix $\mathbf{A}$, resulting in the updated feature vectors $\mathbf{\Tilde{X}} = \textrm{GNN}_\theta (\mathbf{X}, \mathbf{A})$. Finally, an MLP $\phi_o$ considers each of the feature vectors $\mathbf{\Tilde{x}}^{\left(i\right)}$ independently to predict the measurement errors $\hat{e}^{\left(i\right)} = \phi_o (\mathbf{\Tilde{x}}^{\left(i\right)})$. We train this model in a supervised way by computing the $L_2$ loss between the model predictions $\hat{\mathbf{e}}$ and the known labels $\mathbf{e}$ as: 
\begin{equation} 
\label{eq:loss}
L_2(\mathbf{\hat{e}}, \mathbf{e}) = \sum_{i=1}^n \| \hat{e}^{\left(i\right)} - e^{\left(i\right)} \|^2
\end{equation}

\subsection{Measurement Selection}

As shown earlier, our proposed solution does not impose a constraint on the number or the order of pseudo-range measurements in the epoch. Both variants of Cost Function Regulation are applicable, as long as there are more pseudo-range measurements in the epoch than the number of unknowns. When WLS is used to predict the three position coordinates of the receiver and the receiver’s clock bias, at least 4 pseudo-range measurements are required.

\begin{figure}
\begin{minipage}{.48\textwidth}%
\begin{algorithm}[H]
\caption{Proposed measurement selection algorithm}\label{alg:postproc}
\textbf{Initialize:} minimum number of required measurements $n_{req}$\\
\-\hspace{1.5cm} error lower bound $l_b$\\
\-\hspace{1.5cm} error upper bound $u_b$\\
\-\hspace{1.5cm} step size $s$\\
\textbf{Input:} vector of predicted errors $\mathbf{\hat{e}}$\\
\textbf{Output:} vector of selected predictions $\mathbf{\hat{e}_f}$
\begin{algorithmic}
\If{$\Cnt(\mathbf{\hat{e}}) > n_{req}$}
    \While{$\Cnt(l_b \leq \mathbf{\hat{e}} \leq u_b) < n_{req}$}
        \State $u_b \gets u_b + s$
        \If{$u_b \geq \max{(\mathbf{\hat{e}})}$}
            \State $l_b \gets l_b - s$
        \EndIf
    \EndWhile
    \State $\mathbf{\hat{e}_f} \gets \left\{\mathbf{\hat{e}}; \: l_b \leq \mathbf{\hat{e}} \leq u_b \right\}$
\Else{}
    \State $\mathbf{\hat{e}_f} \gets \mathbf{\hat{e}}$
\EndIf
\end{algorithmic}
\end{algorithm}
\end{minipage}%
\hspace{.02\textwidth}
\begin{minipage}{.5\textwidth}%
\centering
\includegraphics[width=1.\linewidth]{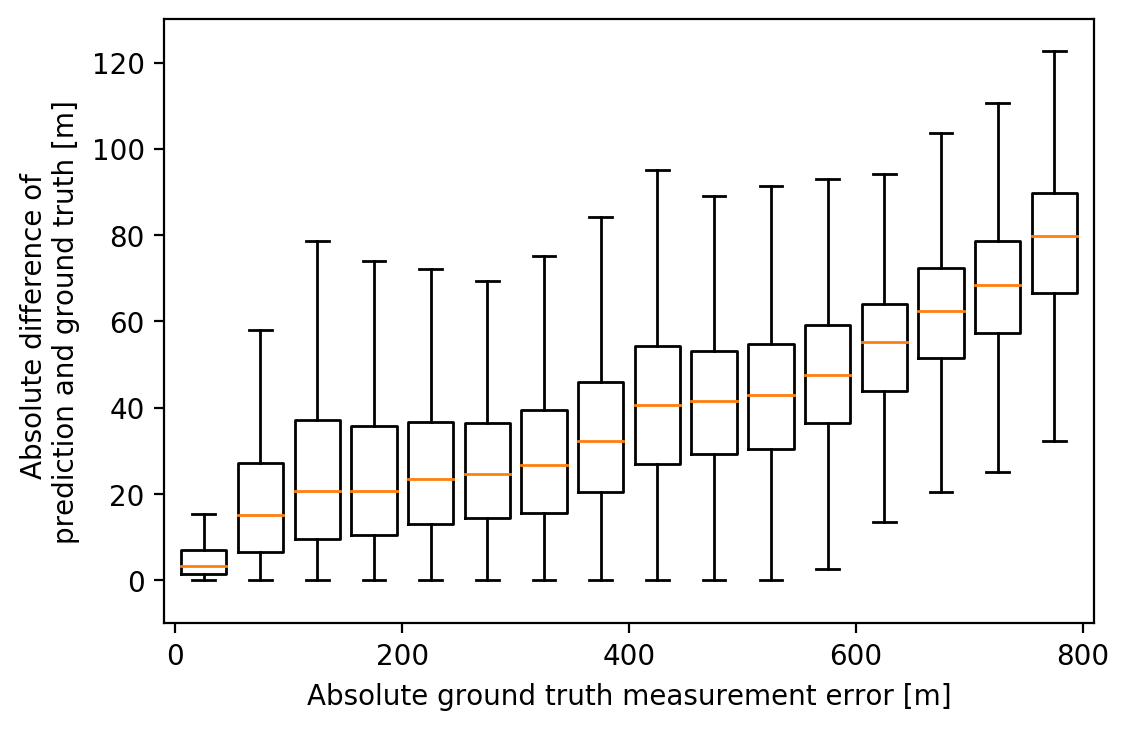}
\caption{Estimator prediction error vs ground truth measurement error. Values of ground truth error are grouped into bins of 50 meters and the Estimator's prediction error distribution is plotted per bin. A clear trend shows that our estimations become less accurate as the measurement error increases.}
\label{fig:trend}
\end{minipage}%
\end{figure}

Moreover, we empirically found that there is an inverse correlation between the absolute values of the measurement errors and the Error Estimator's accuracy. A clear trend is visible in Fig.~\ref{fig:trend} where we plot the absolute prediction error as a function of the absolute measurement error. In this plot, we categorize absolute measurement error into bins of 50 meters. For each bin, the boxplot shows the distribution of the absolute difference between the estimator's prediction and the ground truth. This trend suggests that the GNN's estimations for large errors are less reliable than estimations for lower errors. As our solution has the flexibility to process any subset of size at least 4 from the pseudo-range measurements in the epoch, we design a measurement selection mechanism to discard measurements with large estimated errors.

We would like to select measurements only when we have enough redundancy in the received signals, therefore the selection algorithm needs to adapt to the epoch size. Our proposed measurement selection process works as follows. First, the minimum number of required pseudo-range measurements $n_{req}$ is set. If there are fewer measurements in the epoch, none of them is filtered. If there are more, a lower bound $l_b$ and an upper bound $u_b$  levels of acceptable error are defined. If there are $n_{req}$ measurements with estimated errors within the acceptable interval, the rest of the measurements are filtered out, otherwise, $l_b$ and $u_b$ are gradually relaxed until $n_{req}$ measurements are accepted. The values for $n_{req}$, upper and lower bounds of error are decided by cross-validation on data. Algorithm~\ref{alg:postproc} shows the pseudo-code for our proposed selection algorithm.

This measurement selection algorithm is independent of the error estimation model, adapts to epoch size, and improves localization error by discarding unnecessary pseudo-range measurements.

% ######################################################################################################
% ######################################## EXPERIMENTS ################################################
% ######################################################################################################
\section{Experiments and Results}

\subsection{Dataset}
\label{subsec:exp_dataset}

The training and evaluation data for the model was collected from multiple unconnected drives in four cities from different countries with diverse environmental characteristics and at different times. The dataset consists of roughly 120,000 GNSS epochs with 1-second resolution from a mix of open sky, urban, and dense urban scenarios, and includes multiple GNSS constellations and frequencies. The receiver’s position and time data were obtained from a geodetic-survey grade receiver with a mobile phone form factor antenna and an Inertial Measurement Unit. The pseudo-ranges were measured using the position data and the satellite constellation orbital data, with the receiver’s clock bias measured by a stable external reference clock source.

Each measurement in the dataset is corrected for ionospheric and tropospheric delays using Klobuchar \citep{klobuchar1987ionospheric} and Saastamoinen \citep{bevis1994gps} models respectively. Then, we correct each measurement for the Sagnac effect due to the Earth rotation \citep{bidikar2016sagnac}. The time biases between GNSS constellations are estimated and compensated for, using a preliminary WLS estimation with additional unknowns. By removing these time biases between GNSS constellations and bringing all measurements to the same time offset, measurements from any GNSS constellation can be used in our solution. As a result, the median count of processed measurements per epoch in our solution reaches 60. This relatively large number enables our Measurement Selector block to be activated more often and significantly improve localization. The preliminary WLS is initialized by a location acquired by a cellular network or an externally injected position available on the device. The output of this preliminary WLS provides us with the initial guess for the user location.

We use the cumulative distribution function (CDF) of horizontal error in localization for evaluating methods. Specifically, we use the 50\textsuperscript{th} percentile (median) and the 95\textsuperscript{th} percentile (indicator of difficult situations) of the CDF as evaluation metrics. We divide the dataset into five folds corresponding to non-overlapping geographical regions and report metrics resulting from leave-one-out cross-validation. In this way, we can evaluate our solution on data coming from regions with different environmental characteristics from the ones observed at training time. We also report the standard deviation of the metrics by repeating this process with five random seeds. Initial values of the parameters of the neural network and the loading order of data samples at training time are re-instantiated per seed.

The input features to the Error Estimator include satellite information such as constellation encoding, elevation and azimuth, and signal information such as frequency band, C/N\textsubscript{0}, and average signal power. Estimated residuals at the initial guess location are added as an extra feature to the Error Estimator input. We estimate the receiver clock bias for this initial guess in a way that moves the 10\textsuperscript{th} percentile of the residuals computed at this location to zero. For each fold, we apply standard scaling over the training dataset to all input features and the training label.

\subsection{Experimental Setup}
We implement both the MLP encoder $\phi_e$ and the output MLP $\phi_o$ as a sequence of 5 linear layers followed by Batch Normalization \citep{ioffe2015batch} and Leaky ReLU \citep{maas2013rectifier} layers. The graph neural network $\textrm{GNN}_\theta$ is implemented as 2 consecutive GraphSAGE layers \citep{hamilton2017inductive} to increase the receptive field of the message passing over the graph. We found a larger receptive field to have a detrimental performance which could be explained by the over-smoothing problem on such small graphs \citep{cai2020note, chen2020measuring}. We use weighted mean as the aggregation function to increase the importance of measurements from satellites close to each other. The same setup is used for MP-NN \citep{gilmer2017neural} and GAT \citep{velivckovic2017graph} graph neural networks in our ablation study. The LSTM baseline is implemented following the original architecture described by \citet{zhang2021prediction}. Since visibility labels are not available in our dataset, we only train the LSTM baseline to predict the measurement error.

We train all models with a batch size of 32 over 10k iterations. We use Adam \citep{kingma2014adam} optimizer with a weight decay of $1\text{e-}3$, an initial learning rate of $1\text{e-}3$, and an exponential decay rate of $0.8$ every 1500 iterations.

\begin{figure}[h]
\centering
\includegraphics[width=0.7\textwidth]{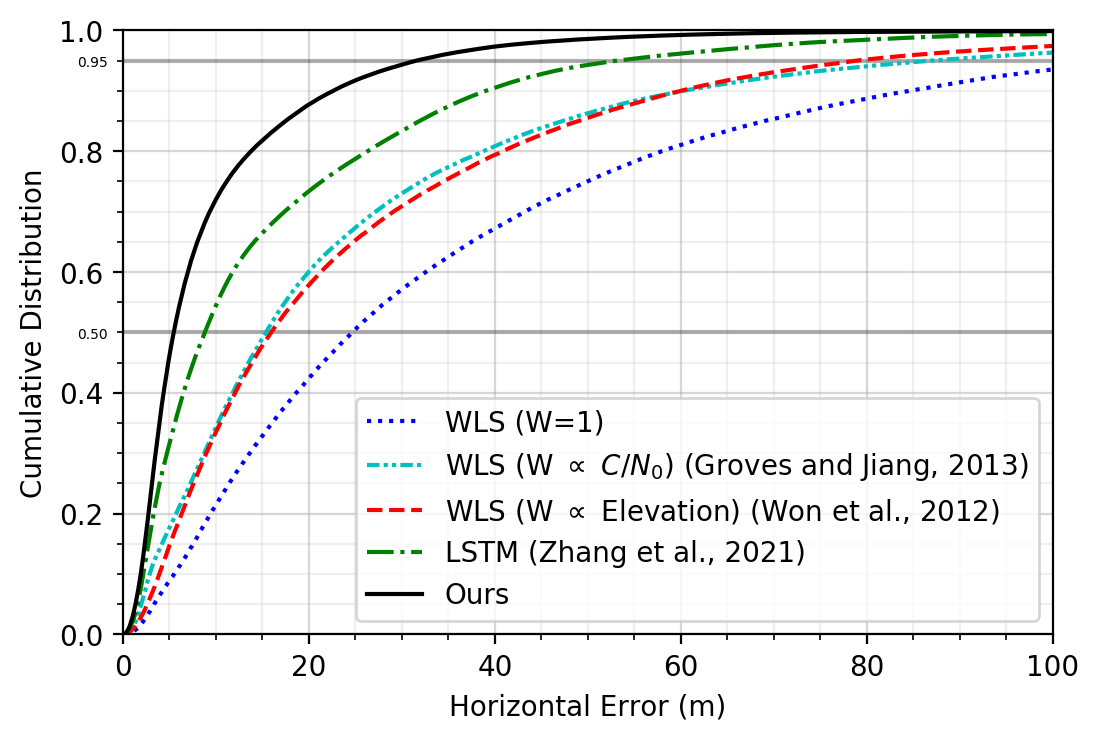}
\caption{Cumulative distribution of horizontal localization error from our proposed solution compared against several baselines.}
\label{fig:experiments_main}
\end{figure}

\subsection{Experiments}
\label{subsec:results}

We compare our end-to-end solution against multiple baselines by visualizing the cumulative distribution of the horizontal localization error in Fig.~\ref{fig:experiments_main}. The first baseline is the least squares method. In this approach, the measurement errors are not estimated and all the weights in WLS are set to one. We also compare against two heuristic approaches. The first approach is C/N\textsubscript{0}-based weighting \citep{groves2013height} where predefined constants are used to map from C/N\textsubscript{0} values to weights for WLS. In the second approach, proposed by \citet{won2012weighted}, the relation between satellite elevations and the measurement errors in the training data is modeled with an exponential function which is used to estimate the optimal weights for WLS at test time. Finally, we compare against a recent deep learning method for measurement error estimation \citep{zhang2021prediction}. In this method, an LSTM-based neural network is employed to obtain a global representation from all measurements in each epoch. This global representation is combined with individual measurement features and processed with an MLP to estimate measurement errors.

\begin{table}[tb]
% increase table row spacing, adjust to taste
\renewcommand{\arraystretch}{1.3}
\caption{Comparison with existing methods based on heuristic and neural network. Our method shows an improvement in horizontal error at the 95\textsuperscript{th} and 50\textsuperscript{th} percentiles. The relative improvement obtained with our method against each baseline is also reported. For methods relying on deep learning models, we report mean and standard deviation over 5 seeds.}
\label{tab:maintab}
\centering

\begin{tabular}{l|cc|cc}
\toprule
 & \multicolumn{2}{c|}{HE @ 95\textsuperscript{th}} & \multicolumn{2}{c}{HE @ 50\textsuperscript{th}} \\
 & Score (m) & \begin{tabular}{@{}c@{}}Improvement \\ with ours \end{tabular} & Score (m) & \begin{tabular}{@{}c@{}}Improvement \\ with ours \end{tabular} \\
\midrule
WLS (W = 1) & 116.43 & 73\% & 24.89 & 78\% \\
WLS (W $\propto$ $\textrm{C/N}_0$) \citep{groves2013height} & 87.01 & 64\% & 15.38 & 65\% \\
WLS (W $\propto$ Elevation) \citep{won2012weighted} & 78.89 & 60\% & 15.92 & 66\% \\
LSTM \citep{zhang2021prediction} & 53.13 ± 0.19 & 41\% & 8.81 ± 0.09 & 38\% \\
\textbf{Ours} & \textbf{31.55 ± 0.83} & --- & \textbf{5.46 ± 0.14} & --- \\
\bottomrule
\end{tabular}
\end{table}

\begin{figure}[!b]%
    \centering
    \subfloat[\centering Data collection route where our solution has the highest improvement over least squares.]{
        {\includegraphics[width=.45\linewidth]{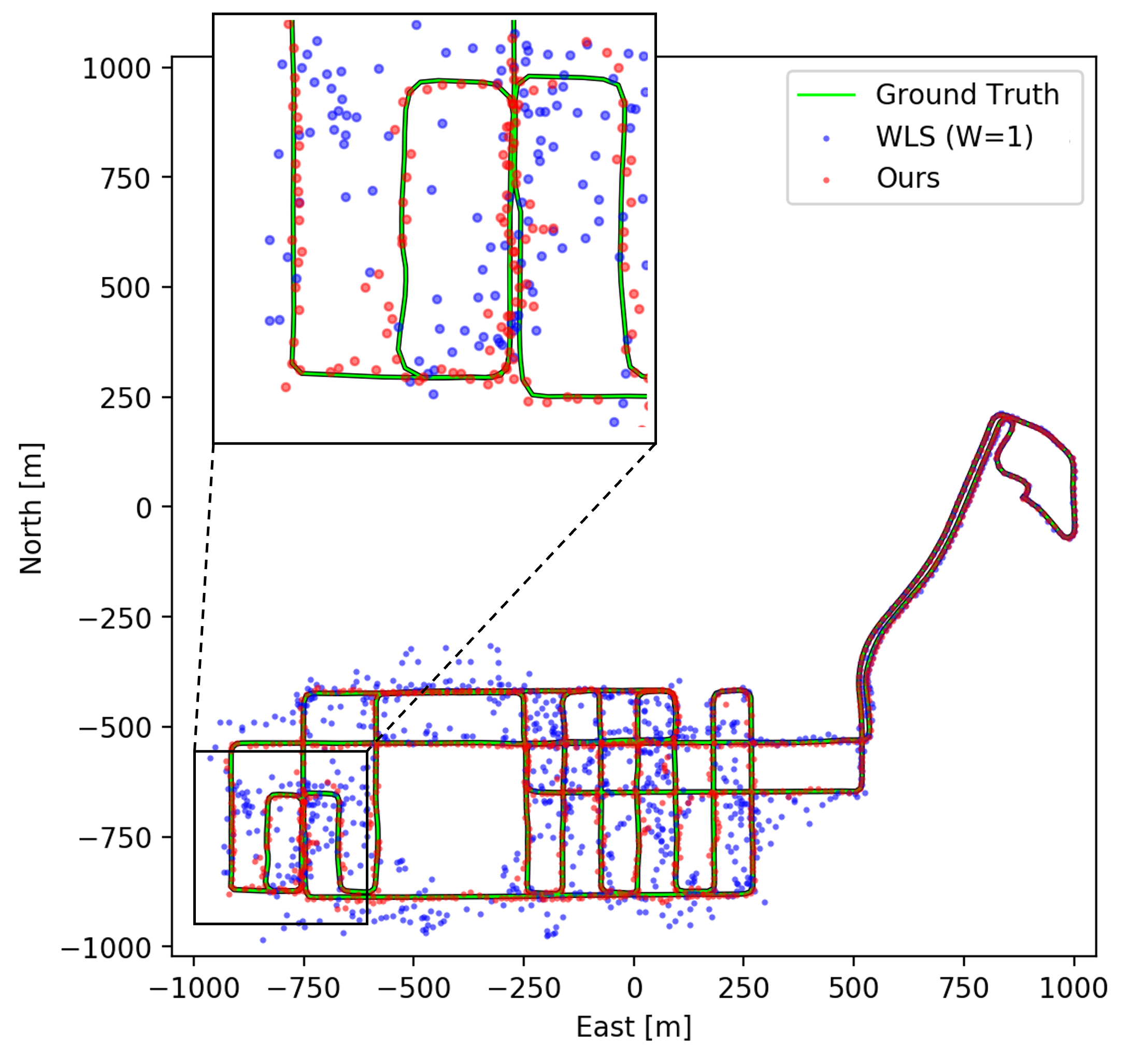} }
        \label{subfig:on_map1}
    }%
    \qquad
    \subfloat[\centering Data collection route where our solution has the lowest improvement over least squares.]{
        {\includegraphics[width=.45\linewidth]{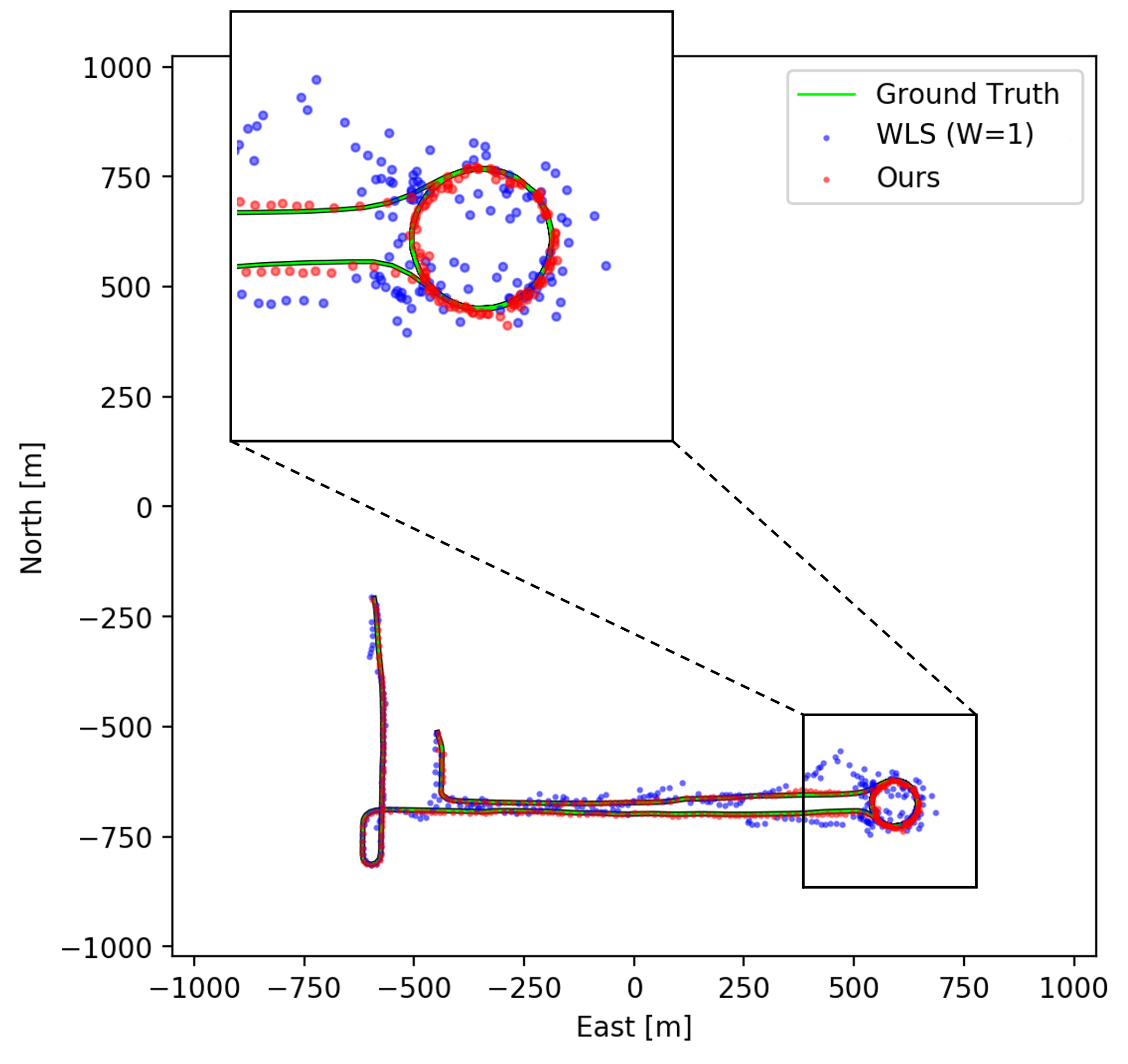} }
        \label{subfig:on_map2}
    }%
    \caption{Predicted location from our solution (red) compared to least squares (blue).}%
    \label{fig:on_map}%
\end{figure}

Table~\ref{tab:maintab} shows the 50\textsuperscript{th} and 95\textsuperscript{th} percentile of the horizontal localization error CDF. Comparing our approach against least squares, we see an improvement of more than 73\%. We observe gains above 60\% over both heuristic measurement weighting approaches. Similarly, our approach consistently outperforms the LSTM-based error estimation approach by over 38\%. To assess the robustness of our method, besides reporting performance variation from seed to seed, we also check for the variation across different folds, where the geographical region of evaluation changes. The standard deviation at the 50\textsuperscript{th} and 95\textsuperscript{th} percentile of horizontal error are about 3 and 8 meters respectively. Therefore, our proposed algorithm generalizes to different scenarios and does not overfit a specific experimental setup. 

In Fig.~\ref{fig:on_map} we show, for two sample data collection routes, the ground truth locations along with the predicted locations using both our method and the least squares algorithm. In route \ref{subfig:on_map1}, our solution has the largest improvement over the least squares method whereas \ref{subfig:on_map2} shows the route where our solution has the smallest improvement. In open sky sections of these routes, where there are many LOS satellites, our Measurement Selector does not filter out many measurements. Since in these regions the measurements have relatively small errors, correcting measurement errors using our Error Estimator does not improve localization remarkably over the baseline least squares algorithm. On the other hand, in downtown areas where there is a limited number of LOS satellites, our solution highly benefits from correcting measurements and drastically outperforms the least squares solution. In general, our location predictions are more stable compared to the baseline as our algorithm is better at reducing extreme measurement errors. We observe similar trends when comparing against other baselines mentioned in Table~\ref{tab:maintab}.

We consider evaluating the performance of the Error Estimator block separately as the rest of the pipeline depends heavily on this component. To evaluate the proposed GNN, we compare ground truth pseudo-range measurement errors against those predicted by our neural network model. Table~\ref{tab:reserr} shows that the mean absolute measurement error is reduced by 78\% when the original measurements are corrected using the GNN’s predictions. Our solution can further improve the corrections from the LSTM-based neural network by 21\%. We visualize absolute measurement error for several data collections in Appendix~\ref{app:viz}. It is clear that in all cases, our algorithm follows the ground truth error more closely.

\begin{table}[tb]
% increase table row spacing, adjust to taste
\renewcommand{\arraystretch}{1.3}
\caption{Mean and median of the absolute error before and after correcting each measurement with predicted values from Error Estimator. The standard deviations are over 5 seeds.}
\label{tab:reserr}
\centering
\begin{tabular}{l|c|c}
\toprule
 % & \begin{tabular}{@{}c@{}}Absolute error\\mean (m) \end{tabular} & \begin{tabular}{@{}c@{}}Absolute error\\median (m) \end{tabular} \\
  & Absolute error mean (m)  & Absolute error median (m)  \\
\midrule
Without correction & 33.65 & 8.05 \\
LSTM \citep{zhang2021prediction} & 9.31 ± 0.18 & 4.75 ± 0.17 \\
\textbf{Ours} & 7.65 ± 0.70 & 4.20 ± 0.39 \\
\bottomrule
\end{tabular}
\end{table}

\begin{figure}[!b]
\centering
\includegraphics[width=0.7\textwidth]{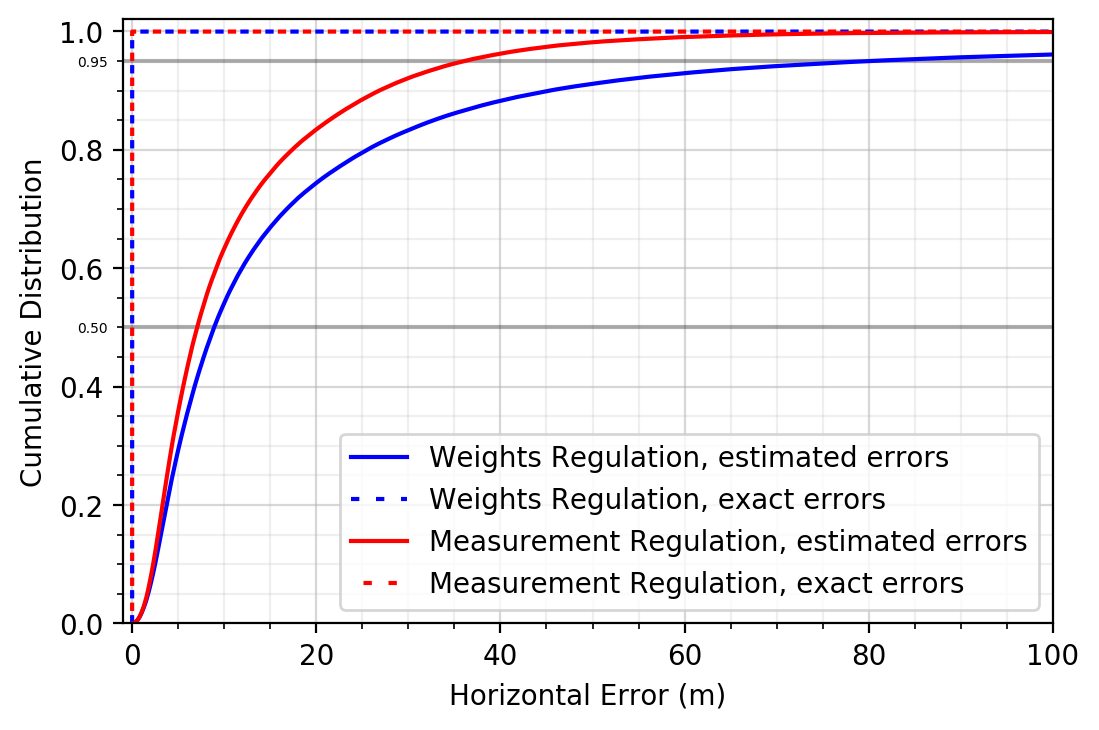}
\caption{Evaluation of the two variants of our solution on real data with estimated and exact measurement errors.}
\label{fig:experiments_wio}
\end{figure}

In Fig.~\ref{fig:experiments_wio}, we include a comparison between the two variants of our Cost Function Regulator: Weight Regulation (blue curves) and Measurement Regulation (red curves). The dashed lines show the cumulative horizontal error distribution when we have access to perfect estimations of the measurement error. This empirically shows that with an improved estimation of the measurement errors, the position estimation error can be reduced to zero. When evaluating them on measurement error estimates obtained from our GNN, however, regulating measurements leads to superior performance. We hypothesize Weight Regulation is less robust to inaccuracies in measurement error estimation because when computing a point in the null space of $\mathbf{H_e^T}$ from Eq.~\ref{eq:hew}, the measurement errors $e^{\left(i\right)}$ appear in the denominator of a fraction. This can make Weight Regulation more sensitive to deviations of estimates $\hat{e}^{\left(i\right)}$ from the truth values $e^{\left(i\right)}$.

\subsection{Ablation Studies}
\label{subsec:ablation}

In this section, we investigate the impact of each component in our pipeline on the horizontal error separately by removing or replacing that component. Table~\ref{tab:ablation} demonstrates the impact of choosing different aggregation methods for the GNN. In our approach, we use the GraphSAGE layer \citep{hamilton2017inductive}, where a weighted adjacency matrix is used to update each measurement's features with the weighted sum of neighboring nodes. We compare our chosen aggregation method with the Graph Attention Network (GAT) layer \citep{velivckovic2017graph} and the Message Passing Neural Network (MP-NN) layer \citep{gilmer2017neural}. In GAT, a self-attention layer is trained to compare features from each pair of measurements and predict weights for the weighted aggregation. In MP-NN, the weights are obtained from an MLP which takes as input the features from two measurements and the adjacency value capturing their angular distance. We can see that the GraphSAGE aggregation outperforms GAT and MP-NN by about 5 and 11 meters respectively at the 95\textsuperscript{th} percentile of horizontal error. We conclude that the simplest aggregation method (GraphSAGE) is enough to learn measurement relations in the relatively small graphs describing GNSS epochs. We hypothesize that more expressive aggregation methods might be learning relations across measurement features that might not translate to unseen test environments. Finally, we include the case in which the measurement encodings in each node are not propagated through the graph. The inferior results of this case confirm that our model benefits from propagating the information across adjacent satellites in the same region of the observable hemisphere.

\begin{table}[!tb]
% increase table row spacing, adjust to taste
\renewcommand{\arraystretch}{1.3}
\caption{Ablation results for the neural network choice for the Error Estimation. %%%and for the impact of removing the Cost Function Regulator and Measurement Selector components in our pipeline%%%
Mean and standard deviation are computed over 5 seeds.}
\label{tab:ablation}
\centering
\begin{tabular}{l|c|c}
\toprule
 & HE @ 95\textsuperscript{th} (m) & HE @ 50\textsuperscript{th} (m) \\
\midrule
Aggregation GraphSAGE (Ours) & 31.55 ± 0.83 & 5.46 ± 0.14 \\
Aggregation GAT & 37.00 ± 1.53 & 6.04 ± 0.46 \\
Aggregation MP-NN & 42.59 ± 0.37 & 6.62 ± 0.11 \\
%Aggregation LSTM (optional) & --- & --- \\
No Aggregation & 40.80 ± 0.27 & 6.38 ± 0.03 \\
%\midrule
\bottomrule
\end{tabular}
\end{table}

Furthermore, we include studies on the impact of the rest of the components in the pipeline. We empirically observe that excluding the Cost Function Regulator block while still keeping the Measurement Selector does not deteriorate the results. This implies that our GNN estimator is not accurate enough in estimating errors for the measurements that pass through our selection mechanism. Error estimation accuracy for these measurements can be improved by fusion of temporal information or additional sensory data such as acceleration from an Inertial Measurement Unit (as shown in Fig.~\ref{fig:experiments_wio}, improvements in error estimation would lead to improvement in positioning accuracy of our solution, unlike the approaches that use heuristics for model supervision). Additionally, a system that solely relies on the selection of measurements would be impractical for real-world applications, since its use would exclusively be limited to epochs with an abundance of measurements, which might not be the case in NLOS-heavy scenarios or with a receiver supporting fewer GNSS constellations.

On the other hand, when we exclude our proposed Measurement Selector block from the pipeline, while still keeping the Cost Function Regulator, we observe that the 95\textsuperscript{th} percentile of the horizontal error CDF degrades from 31 meters to 36 meters. While this shows the importance of removing noisy measurements from the set of available signals, comparing these results against the baselines in Table~\ref{tab:maintab} highlights the effectiveness of Cost Function Regularization.

% ######################################################################################################
% ######################################## CONCLUSION ##################################################
% ######################################################################################################
\section{Conclusion}

In this work, we proposed an innovative end-to-end solution for GNSS positioning based on three components. First, we showed that by regulating the cost function, one can force a location cost optimizer to converge to the truth location. We introduced an analytical method to regulate the cost function using GNSS measurement errors. Second, we showed how these measurement errors can be accurately estimated by designing an expressive GNN to process the GNSS measurements in an epoch. Third, we laid out a measurement selection process to find a robust subset of GNSS measurements in an epoch based on the GNN estimates. We evaluated the performance of the proposed solution on real-world data and showed that it outperforms classical solutions as well as recent neural network-based ones.

\clearpage
%\section*{ACKNOWLEDGEMENTS}
%Any acknowledgements should appear just before the references section at the END of the paper.

% the apacite bibliography style matches the ION bibliography style guidelines.
\bibliographystyle{apalike}
\bibliography{main.bib}

% ######################################################################################################
% ######################################## APPENDIX ####################################################
% ######################################################################################################

\newpage
\begin{appendices}
\section{Measurement Error Visualization}
\label{app:viz}

We compare measurement error predictions from our GNN model with those from the LSTM-based model proposed by \citet{zhang2021prediction}. In Fig.~\ref{fig:meas_err_seq}, we show the absolute difference between ground truth measurement error and model prediction averaged per epoch for several data collection routes. While in some scenarios such as Route A, the performances from the two models are aligned, in most other cases our proposed GNN model outperforms the LSTM-based model (Route C and D). Furthermore, our proposed model is more robust and reduces the average measurement error in epochs with low average error as well as those with high average error. This is not the case for the LSTM-based model using which the average epoch error may increase occasionally as the example in Route B shows.

\renewcommand{\thefigure}{A\arabic{figure}}
\setcounter{figure}{0}

\begin{figure}[!htb]
\centering
\includegraphics[width=1.0\textwidth]{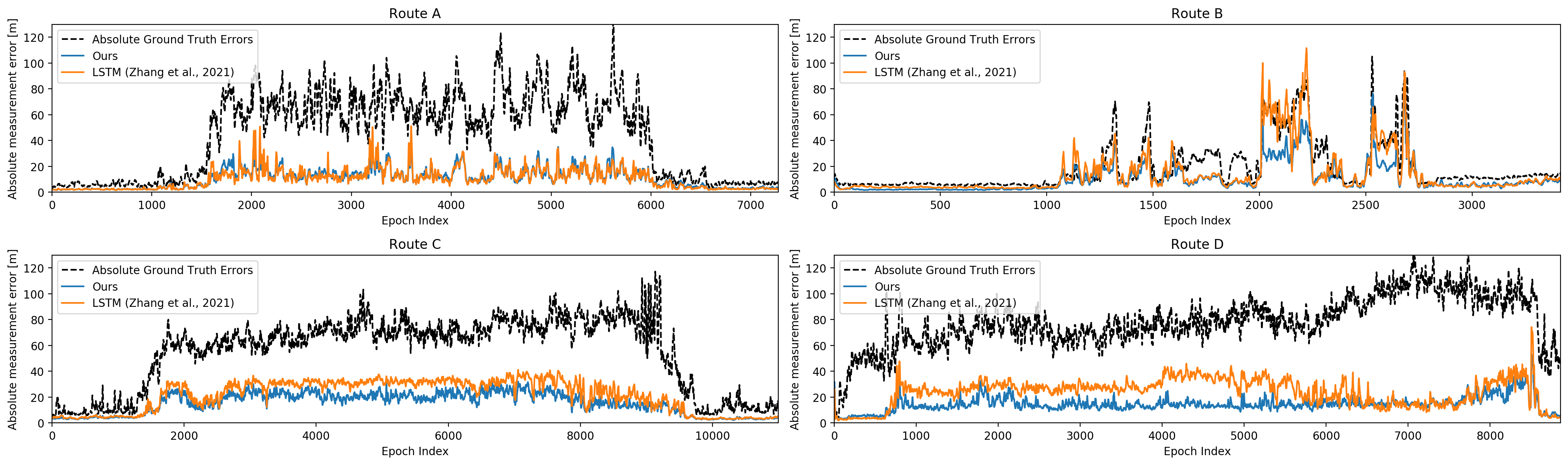}
\caption{Average absolute difference between ground truth measurement errors and model predictions. The average absolute value of ground truth measurement errors is shown in black.}
\label{fig:meas_err_seq}
\end{figure}

\end{appendices}

\end{document}